\documentclass{article}
\usepackage{opcad_meta_style}
\usepackage[T1]{fontenc}
\usepackage{graphicx}
\usepackage{amsmath,amssymb}
\usepackage{booktabs,multirow,array,tabularx}
\usepackage{lmodern}
\definecolor{opcadrow}{gray}{0.92}
\usepackage{placeins}
\usepackage{needspace}
\usepackage{adjustbox}
\usepackage{hyperref}
\usepackage{url}
\hypersetup{colorlinks=true,linkcolor=black,urlcolor=black,citecolor=opcadred,pdftitle={OP-CAD: On-Policy Clean-Audio Distillation for Robust Audio-Visual Reasoning},pdfauthor={Xingming Shui, Dapeng Chen, Bowei Liu, Jingqi Tian, Minfu Li, Kun Yi, Jiapeng Hong, Yansong Tang}}
\title{OP-CAD: On-Policy Clean-Audio Distillation for Robust Audio-Visual Reasoning}
\author{\textbf{Xingming Shui$^{1}$ \quad Dapeng Chen$^{2}$ \quad Bowei Liu$^{1}$ \quad Jingqi Tian$^{1}$}\\[-0.1ex]
\textbf{Minfu Li$^{2}$ \quad Kun Yi$^{2}$ \quad Jiapeng Hong$^{2}$ \quad Yansong Tang$^{1,*}$}\\[0.4ex]
{\small $^{1}$Tsinghua Shenzhen International Graduate School, Tsinghua University}\\[-0.1ex]
{\small $^{2}$Independent Researcher \quad $^{*}$Corresponding author}}
\begin{document}
\maketitle
\thispagestyle{fancy}
\begin{abstract}
Omni-modal large language models deployed in real-world environments encounter external noise that can interfere with their perception and understanding of multimodal inputs. We study their robustness in audio-visual understanding, focusing on question answering under environmental noise and competing speech. The challenge is to resist acoustic interference while preserving useful audio evidence. On-policy distillation provides dense teacher feedback on student-generated responses, but uniform token weighting does not explicitly prioritize positions affected by acoustic interference. We introduce OP-CAD (On-Policy Clean-Audio Distillation), a curriculum-based privileged self-distillation framework for robust audio-visual understanding. Training progresses from mild to severe environmental noise and competing speech, with selective token-level supervision at each stage. The student generates responses from corrupted audio-visual input, while a frozen teacher uses clean audio and the verified answer to supervise the same response prefixes. To allocate this supervision, OP-CAD compares teacher predictions under clean, corrupted, and visual-only contexts without revealing the answer. These matched comparisons measure sensitivity to audio removal and corruption; a bounded weighting rule emphasizes positions identified by either signal while retaining supervision throughout the response. OP-CAD outperforms the compared methods across all evaluated noise conditions. Paired analyses further show improved preservation of clean-correct answers under strong interference, with no observed aggregate clean-accuracy penalty. These results demonstrate the value of directing clean-teacher supervision toward acoustically sensitive predictions for robust audio-visual reasoning.
\end{abstract}
\section{Introduction}

\begin{figure}[t]
    \centering
    \includegraphics[width=0.49\linewidth]{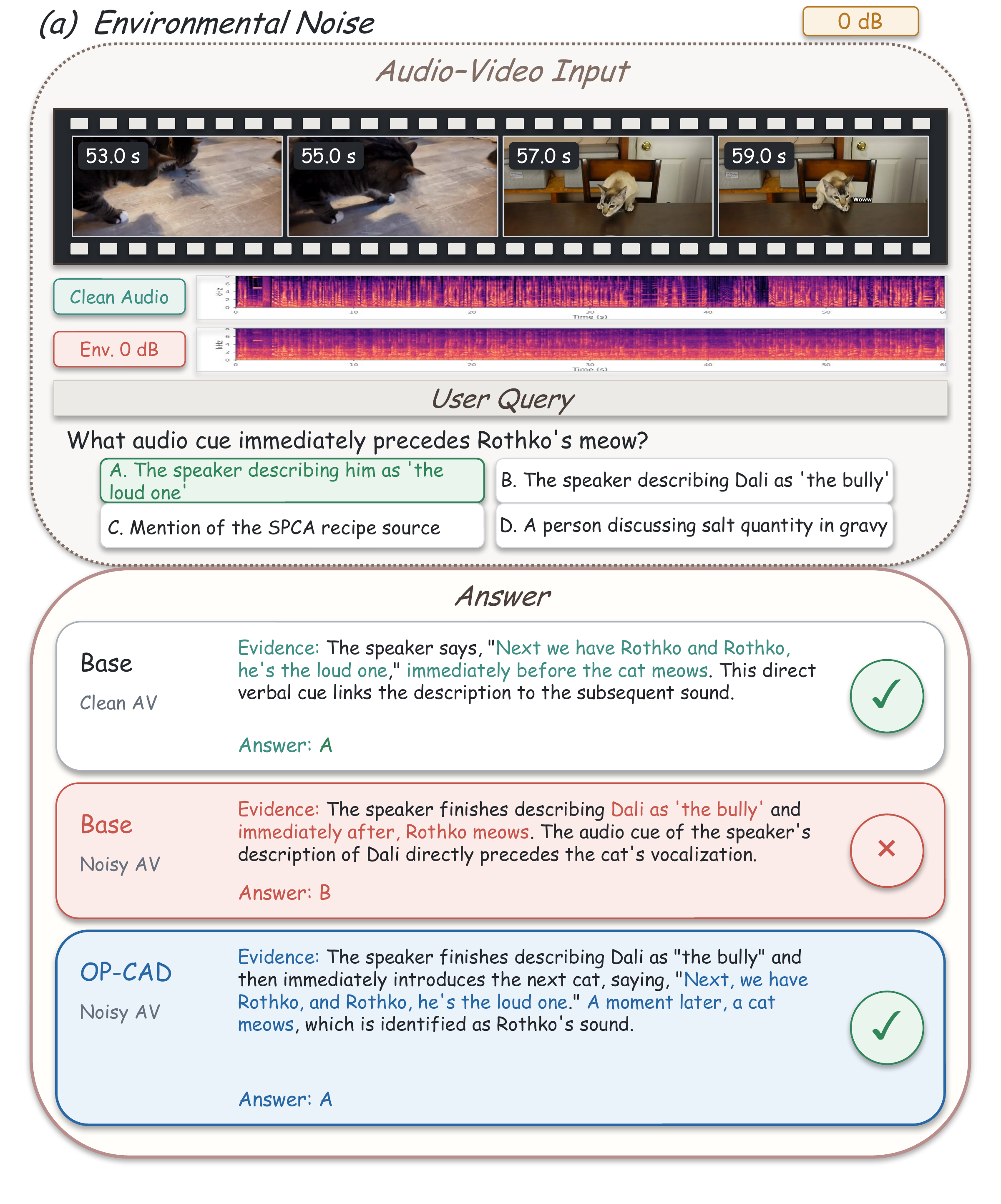}\hspace{0.01\linewidth}
    \includegraphics[width=0.49\linewidth]{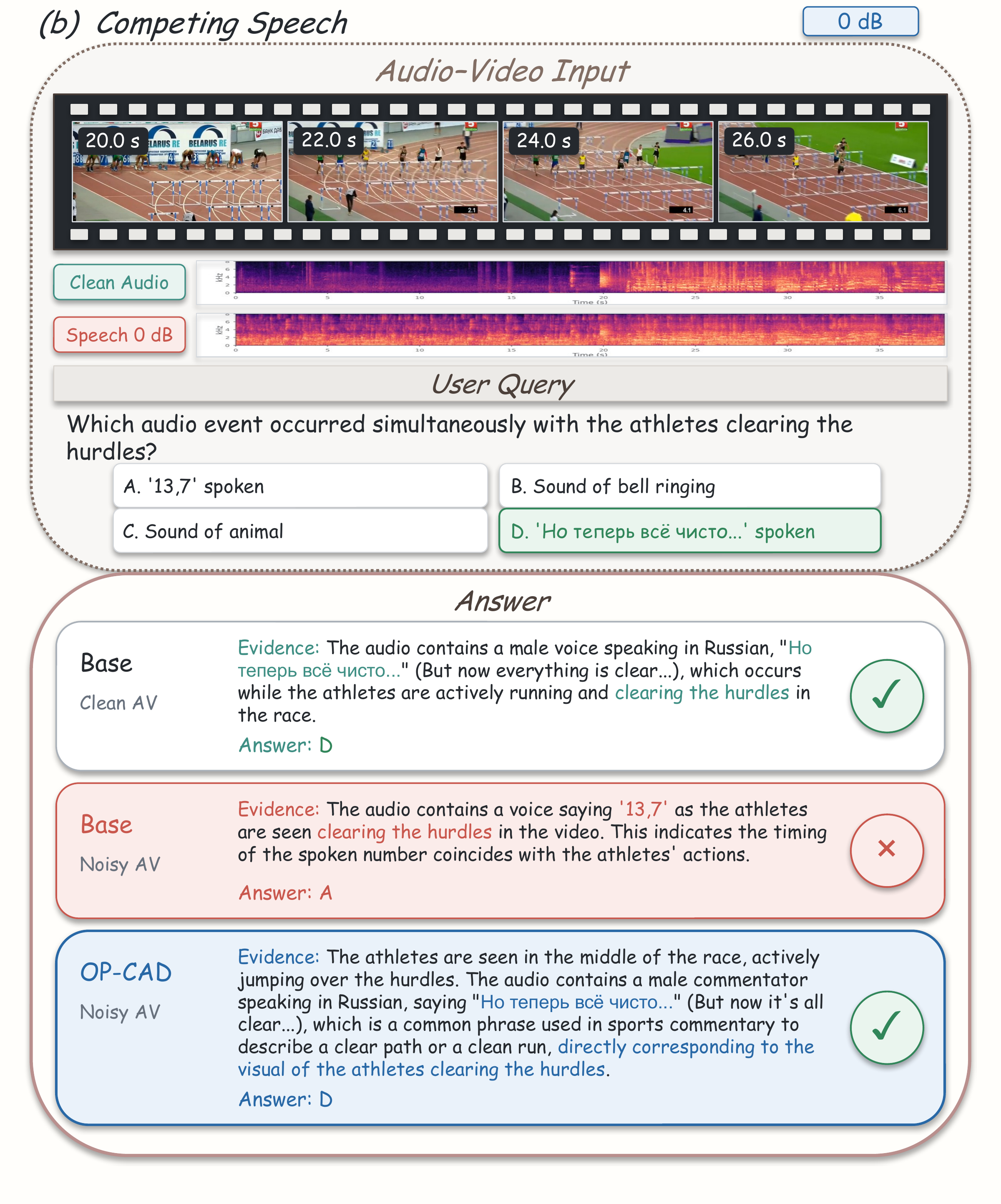}
    \caption{Prediction examples at 0 dB: (a) environmental noise; (b) competing speech.}
    \label{fig:qualitative-recovery}
\end{figure}

Omni-modal large language models jointly interpret visual and acoustic information, providing a perceptual foundation for embodied agents and smartphone assistants that interact with their surroundings \citep{vita,qwen25omni,qwen3omni}. Yet real-world audio often contains environmental noise or background speech. Models that correctly understand a scene in quiet conditions may be misled by irrelevant sounds in noisy environments. Acoustic interference directly affects audio without directly changing the accompanying video; visual information can therefore still provide complementary cues. This motivates using the available visual context to support audio-visual understanding when audio is corrupted. We take audio-visual question answering as our starting point and study how to train omni-modal models to make correct judgments from the available audio-visual evidence under acoustic interference. Figure~\ref{fig:qualitative-recovery} illustrates the problem and our goal: with the video and question unchanged, interference turns the Base model's correct answer into an incorrect one, whereas OP-CAD answers correctly from the corrupted audio-visual input.

Recent work uses privileged information and on-policy self-distillation to improve audio-conditioned understanding and reasoning. CORD uses a text-conditioned teacher to guide audio reasoning \citep{hu2026cord}, while EchoDistill uses a clean-audio teacher to supervise student generation under noisy audio \citep{lin2026echodistill}. These studies demonstrate the value of richer teacher information but focus on audio-conditioned tasks. Unlike audio, visual information in real-world environments is not directly affected by these acoustic disturbances. We therefore aim to retain visual information while using privileged information from clean audio-visual input to guide learning under acoustic interference.

Paired clean and corrupted audio-visual inputs make this form of supervision possible during training. The teacher receives clean audio-visual input, the student receives its corrupted counterpart, and both share the same visual information. The teacher thus has a more complete reference. OPSD provides token-level guidance along the response prefixes actually generated by the student, bringing privileged supervision onto the student's trajectories under interference \citep{zhao2026self}. However, OPSD assigns equal supervision coefficients across positions without using prediction sensitivity to audio information or acoustic interference.

We use the information differences between clean audio-visual input and restricted inputs to construct supervision targets and guide their allocation. Relative to corrupted audio-visual input, clean input provides undistorted acoustic evidence; relative to visual-only input, it provides information from the audio modality. We use both comparisons to identify sensitive positions, increasing clean-teacher supervision when either signal is strong while preserving baseline supervision throughout the response. Privileged information thus guides both what the student should learn and where stronger guidance is needed, connecting complete audio-visual observations to learning under interference.

Building on this idea, we introduce On-Policy Clean-Audio Distillation (OP-CAD), a curriculum-based privileged self-distillation framework. The student generates responses from corrupted audio-visual input, and a frozen teacher uses clean audio-visual input and the verified answer to supervise the same prefixes. Prediction differences induced by audio removal and corruption adjust the training weights. Across examples, training follows a curriculum of progressively stronger interference; within each response, the two signals determine the supervision strength at each position. At inference, the student answers directly from audio-visual input. Experiments show that OP-CAD outperforms the compared baseline training methods under both environmental noise and competing speech, with no observed decrease in aggregate clean accuracy. Ablations further demonstrate the effectiveness of sensitivity-based supervision allocation and the mild-to-severe noise curriculum. Paired analyses show that OP-CAD better preserves clean-correct answers under strong interference than OPSD.

Our contributions are threefold:
\begin{itemize}
    \item We propose a privileged self-distillation framework for noise-robust audio-visual QA. With shared visual context, privileged information from clean audio-visual input guides responses generated by a student under corruption.
    \item We design a supervision allocation mechanism based on two audio condition comparisons. Using clean audio-visual input as a common reference, audio removal and corruption identify acoustically sensitive teacher predictions and determine bounded token-level supervision weights.
    \item We incorporate a mild-to-severe noise curriculum into audio-visual privileged self-distillation, combining progressive interference with selective supervision within responses. Under equal training budgets, this schedule exceeds randomly mixed noise levels under both interference types on both benchmarks.
\end{itemize}

\section{Related Work}

\paragraph{Audio-visual understanding and acoustic robustness.}
Multimodal models have progressed from image and text understanding to video and audio-visual understanding \citep{clip,blip2,llava,video_llava,videollama,videollama2}. Related advances include audio-visual segmentation through disentangled audio semantics \citep{tian2026delayed}, adaptive thinking for video reasoning \citep{tian2026adathinkv}, and memory management for ultra-long video understanding \citep{jin2025videomem}. Omni-modal large language models further integrate vision, audio, and language within a unified generative framework \citep{vita,baichuanomni,qwen25omni,qwen3omni}. Benchmarks such as WorldSense, Daily-Omni, and OmniVideoBench advance the evaluation of joint audio-visual perception and reasoning \citep{worldsense,dailyomni,omnivideobench}. Meanwhile, AVHBench, VoiceBench, RSA-Bench, and AVTrustBench examine reliability through cross-modal conflict, acoustic degradation, and modality dependence, showing that understanding of clean input alone does not ensure reliable behavior under challenging inputs \citep{avhbench,voicebench,rsabench,chowdhury2025avtrustbench}. To address unreliable inputs, Watch or Listen adapts fusion to modality reliability \citep{watchlisten}, while Focus Then Listen combines source separation and modality routing for noise-robust audio-language understanding \citep{yin2026focus}. These studies advance robust perception through evaluation, input processing, and modality utilization. We focus on audio-visual QA with corrupted audio and unchanged video, investigating privileged training supervision for acoustic robustness within shared visual context.

\paragraph{Privileged information and on-policy distillation.}
Knowledge distillation guides student learning through teacher predictions \citep{hinton2015distilling}. Learning with privileged information allows teachers to access additional information unavailable to students during training, such as more complete modalities or cleaner inputs \citep{vapnik2009privileged,garcia2018modality,radevski2023multimodal,hu2025visionaudio}. Multimodal distillation methods use modality-specific saliency to weight distillation losses \citep{jin2021msd} or transfer audio-visual representations to a video-only student \citep{chen2021ccl}; structured self-distillation has also been studied for video reasoning \citep{lin2026visd}. For autoregressive models, on-policy distillation supplies teacher guidance on student-generated response prefixes, covering generation states actually visited by the student \citep{agarwal2024onpolicy}. OPSD applies this principle to self-distillation, using a teacher path with privileged information, such as verified answers, to supervise the student path \citep{zhao2026self}. We adopt this training mechanism with clean audio-visual input as a privileged reference for corrupted input, and further study how input-dependent prediction changes can guide token-level supervision allocation.

\paragraph{Audio-conditioned distillation and supervision allocation.}
Recent work applies privileged distillation to audio understanding and reasoning. CORD uses a text-conditioned teacher to guide audio-conditioned generation, combining importance-weighted token-level distillation with sequence-level GRPO to bridge the audio--text reasoning gap \citep{hu2026cord}. EchoDistill addresses noisy audio through clean-teacher consistency signals and audio-aware rewards, guiding more reliable student responses under interference \citep{lin2026echodistill}. These methods explore text and clean audio privileges, respectively, with a focus on audio-conditioned tasks. In adjacent multimodal tasks, OmniVerifier and OmniVerifier-M1 use verification for visual outcomes \citep{zhang2025omniverifier,zhang2026omniverifierm1}, while VidForensics-M1 uses verifiable temporal evidence for AI-generated video detection \citep{liu2026vidforensicsm1}. In contrast, we study audio-visual QA where visual information remains available despite acoustic interference. Retaining shared visual context, OP-CAD uses clean audio-visual input as a reference to measure teacher prediction changes caused by audio removal and corruption. These changes determine token-level supervision allocation, allowing privileged information from clean inputs to provide both learning targets and supervision strengths. We compare against SFT, GRPO, and uniform OPSD using the same backbone, data, and training protocol to evaluate this allocation mechanism.

\section{Method}
\label{sec:method}

\begin{figure}[!ht]
    \centering
    \includegraphics[width=0.98\textwidth]{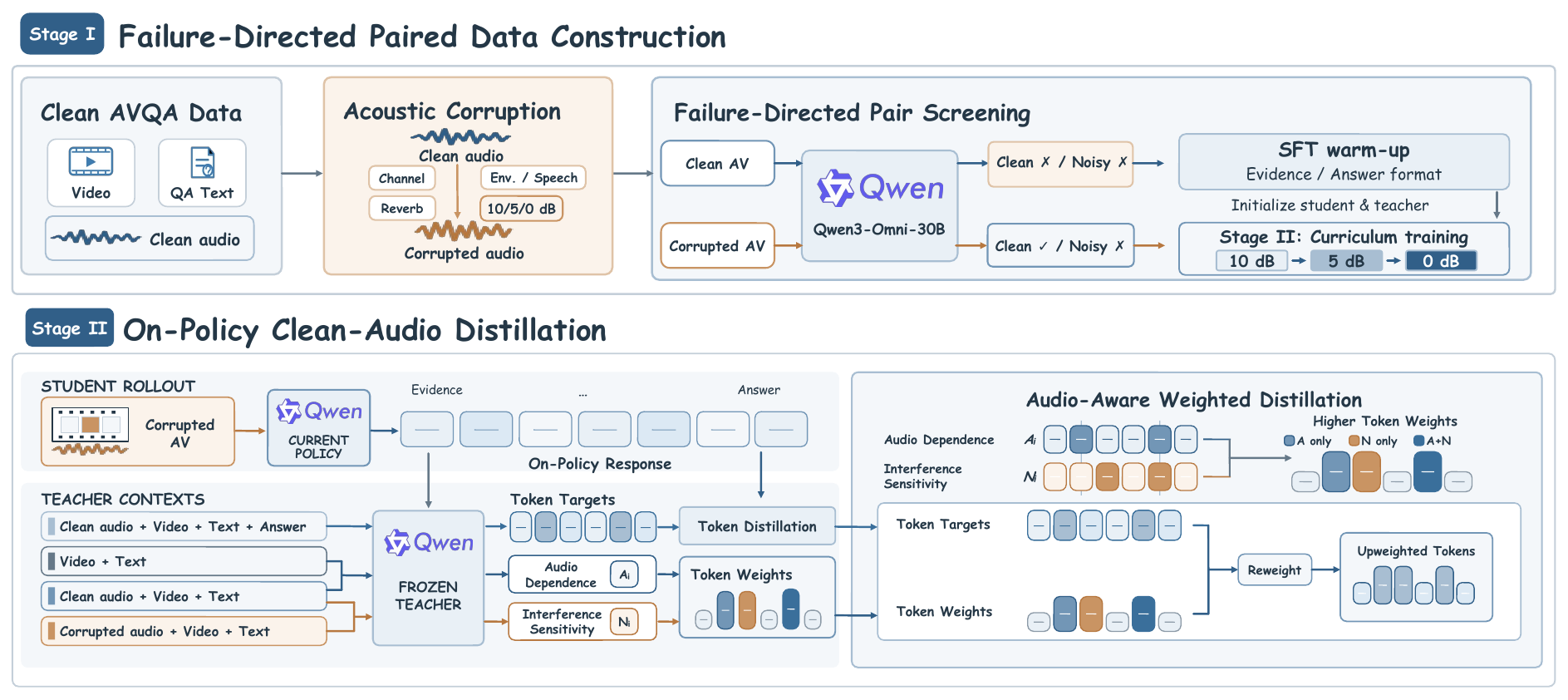}
    \vspace{-2pt}
    \caption{Overview of the OP-CAD training framework.}
    \label{fig:opcad-overview}
\vspace{-8pt}
\end{figure}

OP-CAD uses privileged information from clean audio-visual input to guide audio-visual QA under acoustic interference. The framework has three components. First, we construct clean and corrupted audio versions and select training data through paired screening. Second, supervised warm-up initializes the student and frozen teacher, followed by a curriculum of progressively stronger acoustic interference. Finally, the teacher supervises the student's actual response prefixes, with prediction changes caused by audio removal and corruption determining supervision strength at each position. Figure~\ref{fig:opcad-overview} presents the framework.

\subsection{Paired Audio-Visual Data Construction}
\label{sec:data}

To learn robustness to acoustic interference, we construct paired clean and corrupted audio-visual examples. Candidate questions come from WorldSense \citep{worldsense}, OmniBench \citep{omnibench}, and the AVQA subset of OmniInstruct-v1 \citep{omniinstructv1}. For each example, we modify only the audio data by applying audio degradation.

Specifically, we apply a measured room impulse response, simulate an 8-kHz $\mu$-law channel, and mix environmental noise or competing speech into the clean audio to obtain corrupted audio at SNRs of 10, 5, and 0 dB. Environmental noise comes from MUSAN, DNS Challenge, and NOISEX-92; competing speech comes from VoxCeleb2 \citep{musan,dnschallenge,noisex92,voxceleb2}. Training and evaluation use disjoint interference recordings and room impulse responses, with disjoint speakers for competing speech.

\label{sec:data-routing}
We use the base model $M_0$ to answer the same question under clean and corrupted inputs, partitioning data by the paired predictions:
\begin{itemize}
\item Clean-correct, corrupted-incorrect examples enter the robustness training set $\mathcal D_{\mathrm{rob}}$, focusing training on examples affected by acoustic interference.
\item Examples answered incorrectly under both conditions enter the supervised warm-up set $\mathcal D_{\mathrm{warm}}$ to learn structured evidence and answer outputs.
\item Examples answered correctly under corruption are excluded from both sets, regardless of correctness on clean input.
\end{itemize}

\subsection{Supervised Warm-Up and Noise Curriculum}
\label{sec:warmup-curriculum}

We first fine-tune the base model on $\mathcal D_{\mathrm{warm}}$ to learn the following response structure:
\begin{flushleft}
\texttt{Evidence: <audio-visual evidence supporting the answer>}\\
\texttt{Answer: <one valid option letter>}
\end{flushleft}
Each supervised response pairs evidence generated by Gemini 3 Flash \citep{gemini3flash} with the dataset's verified answer. All generated warm-up evidence underwent human review to check whether the audio-visual content supported the explanation. The resulting model $M_{\mathrm{warm}}$ initializes both the student $S_{\theta_0}$ and teacher $T$. During subsequent robustness training, the student is updated while the teacher remains frozen. The base model $M_0$ used for data screening and the teacher $T$ used for supervision thus serve different roles.

\label{sec:curriculum}
Robustness training follows a $10\!\rightarrow\!5\!\rightarrow\!0$ dB curriculum of increasing acoustic interference \citep{bengio2009curriculum}. We partition $\mathcal D_{\mathrm{rob}}$ into three SNR stages, each containing environmental noise and competing speech examples at its designated level and trained for one epoch. The student generates new responses with its current policy; the teacher remains fixed. The curriculum organizes difficulty across examples, while token weighting organizes supervision within each response, determining the training order and the strength of detailed teacher guidance, respectively.

\subsection{On-Policy Clean-Teacher Supervision}

For each training example, the student receives video $v$, corrupted audio $a^n$, question $q$, and options $\mathcal C$, denoted by $x^n=(v,a^n,q,\mathcal C)$. Replacing the audio with its clean version $a^c$ gives $x^c$; removing audio gives the visual-only input $x^v$. Adding the verified answer $y$ to $x^c$ produces the privileged teacher input $x^{c,+}$.

These inputs serve distinct purposes: $x^{c,+}$ constructs the distillation target, whereas $x^c$, $x^n$, and $x^v$ determine supervision allocation. The latter three exclude the verified answer and share the video, question, and options, so allocation signals reflect prediction differences caused by audio conditions.

The student samples a response $\hat r\sim S_\theta(\cdot\mid x^n)$. Under each input condition, the teacher then predicts at the same prefixes of this response. Comparisons thus correspond to generation states actually visited by the student, without differences in prefixes from independently generated responses.

Let $\mathcal I_x$ index the generated positions of example $x$. At position $i$, the temperature-scaled next-token distributions of the privileged teacher and student are $p_{T,i}^{c,+}=T_\tau(\cdot\mid x^{c,+},\hat r_{<i})$ and $p_{S,i}^{n}=S_{\theta,\tau}(\cdot\mid x^n,\hat r_{<i})$. Following the reverse-KL form of OPSD, the token-level distillation loss is
\begin{equation}
\label{eq:token-opsd}
D_i=\mathrm{KL}\!\left(p_{S,i}^{n}\,\middle\Vert\,p_{T,i}^{c,+}\right).
\end{equation}
The teacher uses clean audio-visual input and the verified answer to provide the learning target, while the student learns to align with it under corrupted input. The verified answer is used only for the privileged target, not for the weight computation below.

\subsection{Supervision Allocation through Audio Condition Comparisons}

\begin{figure}[!htbp]
    \centering
    \includegraphics[width=0.93\textwidth]{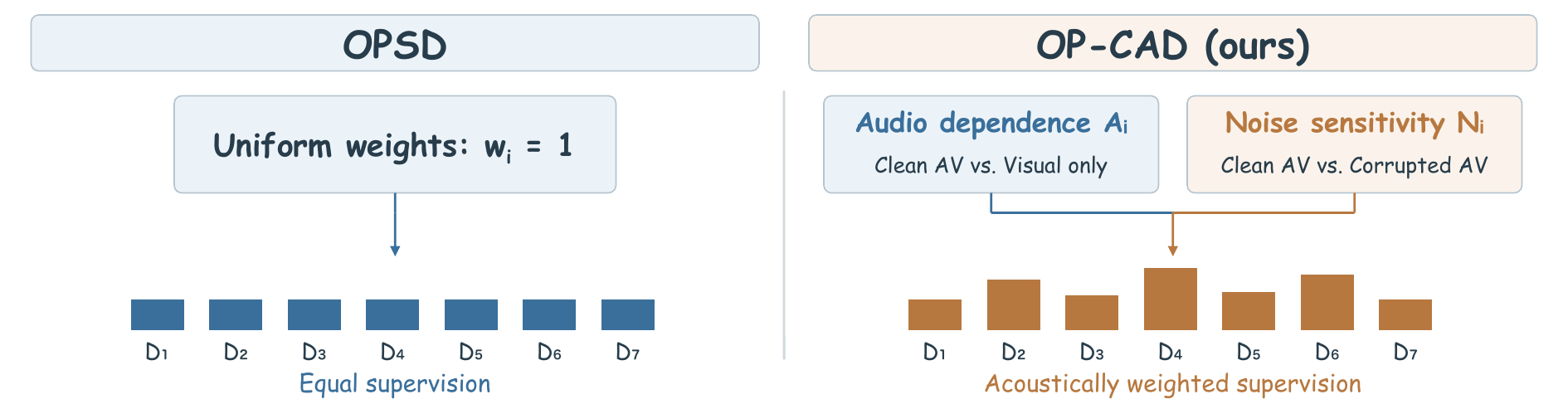}
    \vspace{-2pt}
    \caption{Token-level supervision allocation in OPSD and OP-CAD.}
    \label{fig:intro-objectives}
\vspace{-8pt}
\end{figure}

Uniform OPSD assigns the same weight to every $D_i$. OP-CAD instead uses teacher prediction sensitivity to changes in audio conditions to identify positions requiring stronger supervision. Figure~\ref{fig:intro-objectives} contrasts the two approaches.

For $u\in\{c,n,v\}$, let $p_{T,i}^{u}=T_\tau(\cdot\mid x^u,\hat r_{<i})$ denote the teacher prediction under the corresponding input. Using clean audio-visual input as a common reference, we construct two signals:
\begin{equation}
\label{eq:counterfactual-statistics}
 A_i=\operatorname{JS}(p_{T,i}^{c},p_{T,i}^{v}),
 \qquad
 N_i=\operatorname{JS}(p_{T,i}^{c},p_{T,i}^{n}),
\end{equation}
where $\operatorname{JS}$ denotes Jensen--Shannon divergence. Audio removal sensitivity $A_i$ measures how much the teacher prediction changes when audio is removed while the video and prefix remain fixed. Corruption sensitivity $N_i$ measures the change when clean audio is replaced by corrupted audio. The former captures the prediction difference associated with audio beyond the visual context; the latter captures the difference caused by acoustic interference.

Because the two divergences may have different scales, we normalize each by its maximum within the current response to obtain $\widetilde A_i$ and $\widetilde N_i$. If a signal's response maximum is at most $\epsilon$, its normalized values are set to zero; otherwise, each value is divided by that maximum. The normalized signals lie in $[0,1]$ and represent relative sensitivity within a response. We then take their maximum and construct the supervision weight:
\begin{equation}
\label{eq:allocation-weight}
 s_i=\max(\widetilde A_i,\widetilde N_i),
 \qquad w_i=1+\lambda\,\operatorname{stopgrad}(s_i),\quad \lambda\geq0.
\end{equation}
The baseline weight of 1 preserves supervision at every generated position, while the increment is bounded by $\lambda$, giving $w_i\in[1,1+\lambda]$. Positions sensitive to both interventions do not exceed this bound through signal addition. The $\operatorname{stopgrad}$ operation treats the computed allocation score as constant during backpropagation.

For a mini-batch $\mathcal B$, let $D_{x,i}$ and $w_{x,i}$ denote the distillation loss and supervision weight at position $i$ of example $x$. The final objective is
\begin{equation}
\label{eq:opcad-loss}
\mathcal L_{\mathrm{OP\text{-}CAD}}(\theta)=
\frac{\sum_{x\in\mathcal B}\sum_{i\in\mathcal I_x}w_{x,i}D_{x,i}}
{\sum_{x\in\mathcal B}\sum_{i\in\mathcal I_x}w_{x,i}}.
\end{equation}
Each example contributes one student response, and all teacher conditions score its same prefixes. Sampled prefixes, teacher distributions, and supervision weights are detached; gradients flow only through the student distribution under corrupted input. Normalizing by the total weight preserves relative allocation while keeping the loss scale comparable to uniform OPSD. Setting $\lambda=0$ recovers the uniform OPSD objective.

\newcommand{\scorechange}[3]{\mbox{\strut#1\,\raisebox{0pt}[0pt][0pt]{\scriptsize$#2$#3}}}
\newcommand{\stackchange}[3]{\mbox{\strut#1\kern0.4pt\raisebox{0pt}[0pt][0pt]{\scriptsize$#2$#3}}}
\section{Experiments}
\label{sec:experiments}

\subsection{Experimental Design}

We evaluate all 1,197 questions in Daily-Omni and 1,000 questions in OmniVideoBench under clean audio and environmental noise or competing speech at SNRs of 10, 5, and 0 dB. We compare the original Qwen3-Omni-Instruct model (Base), the SFT warm-up checkpoint, and continued training with SFT, GRPO, OPSD, or OP-CAD. Base and SFT warm-up measure performance before and after supervised initialization. SFT, GRPO, OPSD, and OP-CAD compare robustness training strategies with shared warm-up initialization, data, and noise severity curriculum. SFT uses fixed evidence and answer targets; GRPO \citep{shao2024deepseekmath} samples eight responses per prompt with correctness and format rewards. OPSD \citep{zhao2026self} uses the same frozen teacher and privileged target as OP-CAD but assigns uniform token weights, providing the direct control for supervision allocation.

Training uses rank-64 LoRA \citep{hu2022lora} on the language backbone and multimodal projectors, with frozen audio and visual encoders, a learning rate of $10^{-5}$, and an effective batch size of 32. OPSD and OP-CAD sample one response of up to 128 new tokens per example, using the complete vocabulary with $\tau=1$ and OP-CAD weight strength $\lambda=0.5$. We report deterministic multiple-choice accuracy and the \emph{harm rate}, the fraction of each model's clean-correct answers that become incorrect under corruption.

\begin{table}[!htbp]
\caption{Full test accuracy (\%). Env./Speech: three-SNR averages. Bold: best result. Arrows: changes from SFT warm-up in percentage points, using displayed scores.}
\label{tab:main-results}
\centering\small
\setlength{\tabcolsep}{2pt}
\renewcommand{\arraystretch}{1.12}
\begin{tabularx}{\linewidth}{l*{6}{>{\raggedleft\arraybackslash}X}}
\toprule
& \multicolumn{3}{c}{Daily-Omni} & \multicolumn{3}{c}{OmniVideoBench} \\
\cmidrule(lr){2-4}\cmidrule(lr){5-7}
Method & Clean & Env. & Speech & Clean & Env. & Speech \\
\midrule
Base & 67.75 & 61.38 & 57.84 & 37.40 & 36.70 & 34.80 \\
\midrule
SFT warm-up & 68.76 & 61.74 & 59.82 & 38.10 & 36.80 & 35.93 \\
+ SFT & \scorechange{68.42}{\downarrow}{0.34} & \scorechange{61.88}{\uparrow}{0.14} & \scorechange{59.12}{\downarrow}{0.70} & \scorechange{39.00}{\uparrow}{0.90} & \scorechange{36.23}{\downarrow}{0.57} & \scorechange{36.20}{\uparrow}{0.27} \\
+ GRPO & \scorechange{68.34}{\downarrow}{0.42} & \scorechange{62.91}{\uparrow}{1.17} & \scorechange{59.93}{\uparrow}{0.11} & \scorechange{41.50}{\uparrow}{3.40} & \scorechange{37.87}{\uparrow}{1.07} & \scorechange{36.87}{\uparrow}{0.94} \\
+ OPSD & \scorechange{69.34}{\uparrow}{0.58} & \scorechange{63.49}{\uparrow}{1.75} & \scorechange{60.62}{\uparrow}{0.80} & \scorechange{41.40}{\uparrow}{3.30} & \scorechange{37.40}{\uparrow}{0.60} & \scorechange{35.13}{\downarrow}{0.80} \\
\rowcolor{opcadrow}
\textbf{+ OP-CAD} & \scorechange{\textbf{70.09}}{\uparrow}{1.33} & \scorechange{\textbf{66.97}}{\uparrow}{5.23} & \scorechange{\textbf{63.55}}{\uparrow}{3.73} & \scorechange{\textbf{41.60}}{\uparrow}{3.50} & \scorechange{\textbf{40.23}}{\uparrow}{3.43} & \scorechange{\textbf{38.93}}{\uparrow}{3.00} \\
\bottomrule
\end{tabularx}
\vspace{-6pt}
\end{table}

\subsection{Robustness Across Acoustic Conditions}

Table~\ref{tab:main-results} separates the effect of supervised initialization from subsequent robustness training. Warm-up improves accuracy on clean and corrupted inputs over Base on both benchmarks, but continued training does not yield uniform gains across strategies and conditions. OP-CAD achieves the highest aggregate accuracy among the listed methods in clean audio and both interference families.

OPSD provides the direct comparison for supervision allocation because it shares OP-CAD's teacher target and training protocol. OP-CAD improves environmental noise and competing speech averages by 3.48 and 2.92 percentage points on Daily-Omni and by 2.83 and 3.80 points on OmniVideoBench. It also exceeds OPSD at every tested SNR. These results show that weighting supervision by acoustic sensitivity improves robustness over uniform distillation with the same teacher target.

\begin{table}[!htb]
\caption{Daily-Omni accuracy (\%) by question type and duration. Arrows: changes from Base within each condition in percentage points, using displayed scores. Bold: best per column.}
\label{tab:daily-category-results}
\centering
{\fontsize{8.25}{10}\selectfont
\setlength{\tabcolsep}{0.60pt}
\renewcommand{\arraystretch}{1.12}
\begin{tabular}{llrrrrrrrrr}
\toprule
\multicolumn{11}{c}{\textit{Daily-Omni}} \\
\multirow{2}{*}{Cond.} & \multirow{2}{*}{Method} & \multicolumn{6}{c}{Question Type} & \multicolumn{2}{c}{Video Duration} & \multirow{2}{*}{Avg.} \\
\cmidrule(lr){3-8}\cmidrule(lr){9-10}
& & AV Align & Comp. & Ctx. Und. & Evt. Seq. & Infer. & Reas. & 30s & 60s & \\
\midrule
\multirow{5}{*}{Clean}
& Base & 58.82 & 77.86 & 60.62 & 60.13 & 81.82 & 81.14 & 68.01 & 67.45 & 67.75 \\
 & + SFT & \stackchange{\textbf{61.34}}{\uparrow}{2.52} & \stackchange{77.10}{\downarrow}{0.76} & \stackchange{64.77}{\uparrow}{4.15} & \stackchange{57.84}{\downarrow}{2.29} & \stackchange{83.77}{\uparrow}{1.95} & \stackchange{80.57}{\downarrow}{0.57} & \stackchange{68.62}{\uparrow}{0.61} & \stackchange{\textbf{68.18}}{\uparrow}{0.73} & \stackchange{68.42}{\uparrow}{0.67} \\
 & + GRPO & \stackchange{59.24}{\uparrow}{0.42} & \stackchange{79.39}{\uparrow}{1.53} & \stackchange{63.73}{\uparrow}{3.11} & \stackchange{59.80}{\downarrow}{0.33} & \stackchange{82.47}{\uparrow}{0.65} & \stackchange{80.00}{\downarrow}{1.14} & \stackchange{70.48}{\uparrow}{2.47} & \stackchange{65.82}{\downarrow}{1.63} & \stackchange{68.34}{\uparrow}{0.59} \\
 & + OPSD & \stackchange{56.72}{\downarrow}{2.10} & \stackchange{77.86}{\leftrightarrow}{0.00} & \stackchange{\textbf{68.91}}{\uparrow}{8.29} & \stackchange{\textbf{60.78}}{\uparrow}{0.65} & \stackchange{84.42}{\uparrow}{2.60} & \stackchange{\textbf{82.29}}{\uparrow}{1.15} & \stackchange{71.41}{\uparrow}{3.40} & \stackchange{66.91}{\downarrow}{0.54} & \stackchange{69.34}{\uparrow}{1.59} \\
\rowcolor{opcadrow}
 & \textbf{+ OP-CAD} & \stackchange{60.50}{\uparrow}{1.68} & \stackchange{\textbf{80.15}}{\uparrow}{2.29} & \stackchange{\textbf{68.91}}{\uparrow}{8.29} & \stackchange{59.80}{\downarrow}{0.33} & \stackchange{\textbf{85.06}}{\uparrow}{3.24} & \stackchange{81.71}{\uparrow}{0.57} & \stackchange{\textbf{72.18}}{\uparrow}{4.17} & \stackchange{67.64}{\uparrow}{0.19} & \stackchange{\textbf{70.09}}{\uparrow}{2.34} \\
\midrule
\multirow{5}{*}{Env.}
& Base & 49.30 & 70.99 & 59.93 & 54.68 & 75.97 & 71.05 & 62.34 & 60.24 & 61.38 \\
 & + SFT & \stackchange{54.06}{\uparrow}{4.76} & \stackchange{69.72}{\downarrow}{1.27} & \stackchange{59.93}{\leftrightarrow}{0.00} & \stackchange{53.81}{\downarrow}{0.87} & \stackchange{75.54}{\downarrow}{0.43} & \stackchange{70.86}{\downarrow}{0.19} & \stackchange{62.44}{\uparrow}{0.10} & \stackchange{61.21}{\uparrow}{0.97} & \stackchange{61.88}{\uparrow}{0.50} \\
 & + GRPO & \stackchange{50.70}{\uparrow}{1.40} & \stackchange{74.81}{\uparrow}{3.82} & \stackchange{60.62}{\uparrow}{0.69} & \stackchange{55.12}{\uparrow}{0.44} & \stackchange{76.62}{\uparrow}{0.65} & \stackchange{74.67}{\uparrow}{3.62} & \stackchange{64.45}{\uparrow}{2.11} & \stackchange{61.09}{\uparrow}{0.85} & \stackchange{62.91}{\uparrow}{1.53} \\
 & + OPSD & \stackchange{51.54}{\uparrow}{2.24} & \stackchange{73.03}{\uparrow}{2.04} & \stackchange{60.79}{\uparrow}{0.86} & \stackchange{55.66}{\uparrow}{0.98} & \stackchange{79.44}{\uparrow}{3.47} & \stackchange{75.24}{\uparrow}{4.19} & \stackchange{65.79}{\uparrow}{3.45} & \stackchange{60.79}{\uparrow}{0.55} & \stackchange{63.49}{\uparrow}{2.11} \\
\rowcolor{opcadrow}
 & \textbf{+ OP-CAD} & \stackchange{\textbf{54.76}}{\uparrow}{5.46} & \stackchange{\textbf{77.86}}{\uparrow}{6.87} & \stackchange{\textbf{65.28}}{\uparrow}{5.35} & \stackchange{\textbf{57.84}}{\uparrow}{3.16} & \stackchange{\textbf{83.12}}{\uparrow}{7.15} & \stackchange{\textbf{79.05}}{\uparrow}{8.00} & \stackchange{\textbf{68.62}}{\uparrow}{6.28} & \stackchange{\textbf{65.03}}{\uparrow}{4.79} & \stackchange{\textbf{66.97}}{\uparrow}{5.59} \\
\midrule
\multirow{5}{*}{Speech}
& Base & 47.34 & 66.92 & 58.03 & 50.22 & 71.43 & 66.48 & 58.22 & 57.39 & 57.84 \\
 & + SFT & \stackchange{48.74}{\uparrow}{1.40} & \stackchange{67.18}{\uparrow}{0.26} & \stackchange{56.82}{\downarrow}{1.21} & \stackchange{53.49}{\uparrow}{3.27} & \stackchange{74.03}{\uparrow}{2.60} & \stackchange{66.48}{\leftrightarrow}{0.00} & \stackchange{59.61}{\uparrow}{1.39} & \stackchange{58.55}{\uparrow}{1.16} & \stackchange{59.12}{\uparrow}{1.28} \\
 & + GRPO & \stackchange{46.50}{\downarrow}{0.84} & \stackchange{72.01}{\uparrow}{5.09} & \stackchange{57.86}{\downarrow}{0.17} & \stackchange{52.83}{\uparrow}{2.61} & \stackchange{74.24}{\uparrow}{2.81} & \stackchange{71.24}{\uparrow}{4.76} & \stackchange{61.15}{\uparrow}{2.93} & \stackchange{58.48}{\uparrow}{1.09} & \stackchange{59.93}{\uparrow}{2.09} \\
 & + OPSD & \stackchange{47.48}{\uparrow}{0.14} & \stackchange{74.05}{\uparrow}{7.13} & \stackchange{56.82}{\downarrow}{1.21} & \stackchange{53.49}{\uparrow}{3.27} & \stackchange{75.11}{\uparrow}{3.68} & \stackchange{72.38}{\uparrow}{5.90} & \stackchange{63.01}{\uparrow}{4.79} & \stackchange{57.82}{\uparrow}{0.43} & \stackchange{60.62}{\uparrow}{2.78} \\
\rowcolor{opcadrow}
 & \textbf{+ OP-CAD} & \stackchange{\textbf{50.42}}{\uparrow}{3.08} & \stackchange{\textbf{75.06}}{\uparrow}{8.14} & \stackchange{\textbf{63.04}}{\uparrow}{5.01} & \stackchange{\textbf{55.77}}{\uparrow}{5.55} & \stackchange{\textbf{77.49}}{\uparrow}{6.06} & \stackchange{\textbf{74.67}}{\uparrow}{8.19} & \stackchange{\textbf{65.33}}{\uparrow}{7.11} & \stackchange{\textbf{61.45}}{\uparrow}{4.06} & \stackchange{\textbf{63.55}}{\uparrow}{5.71} \\
\bottomrule
\end{tabular}
}

\vspace{-6pt}
\end{table}

Tables~\ref{tab:daily-category-results} and~\ref{tab:omnivideo-category-results} examine whether aggregate improvements are concentrated in a particular task or duration. OP-CAD exceeds OPSD in every corrupted-condition category and duration column. The observed gains extend across multiple question types, audio types, and video lengths.

\begin{table}[!htb]

\caption{OmniVideoBench accuracy (\%) by audio type and duration. Arrows: changes from Base within each condition in percentage points, using displayed scores. Bold: best per column.}
\label{tab:omnivideo-category-results}
\centering
{\fontsize{8.25}{10}\selectfont
\setlength{\tabcolsep}{0.60pt}
\renewcommand{\arraystretch}{1.12}
\begin{tabular}{llrrrrrrrr}
\toprule
\multicolumn{10}{c}{\textit{OmniVideoBench}} \\
\multirow{2}{*}{Cond.} & \multirow{2}{*}{Method} & \multicolumn{3}{c}{Audio Type} & \multicolumn{4}{c}{Video Duration} & \multirow{2}{*}{Avg.} \\
\cmidrule(lr){3-5}\cmidrule(lr){6-9}
& & Music & Sound & Speech & (0,1] min & (1,5] min & (5,10] min & (10,30] min & \\
\midrule
\multirow{5}{*}{Clean}
& Base & 30.77 & 32.65 & 39.11 & 49.12 & 35.29 & 36.40 & 33.33 & 37.40 \\
 & + SFT & \stackchange{32.97}{\uparrow}{2.20} & \stackchange{37.41}{\uparrow}{4.76} & \stackchange{40.03}{\uparrow}{0.92} & \stackchange{47.95}{\downarrow}{1.17} & \stackchange{39.41}{\uparrow}{4.12} & \stackchange{35.96}{\downarrow}{0.44} & \stackchange{35.25}{\uparrow}{1.92} & \stackchange{39.00}{\uparrow}{1.60} \\
 & + GRPO & \stackchange{\textbf{36.26}}{\uparrow}{5.49} & \stackchange{36.73}{\uparrow}{4.08} & \stackchange{\textbf{43.04}}{\uparrow}{3.93} & \stackchange{49.12}{\leftrightarrow}{0.00} & \stackchange{40.59}{\uparrow}{5.30} & \stackchange{39.47}{\uparrow}{3.07} & \stackchange{\textbf{39.46}}{\uparrow}{6.13} & \stackchange{41.50}{\uparrow}{4.10} \\
 & + OPSD & \stackchange{28.57}{\downarrow}{2.20} & \stackchange{\textbf{40.82}}{\uparrow}{8.17} & \stackchange{\textbf{43.04}}{\uparrow}{3.93} & \stackchange{48.54}{\downarrow}{0.58} & \stackchange{42.06}{\uparrow}{6.77} & \stackchange{\textbf{40.79}}{\uparrow}{4.39} & \stackchange{36.40}{\uparrow}{3.07} & \stackchange{41.40}{\uparrow}{4.00} \\
\rowcolor{opcadrow}
 & \textbf{+ OP-CAD} & \stackchange{32.97}{\uparrow}{2.20} & \stackchange{39.46}{\uparrow}{6.81} & \stackchange{\textbf{43.04}}{\uparrow}{3.93} & \stackchange{\textbf{49.71}}{\uparrow}{0.59} & \stackchange{\textbf{43.53}}{\uparrow}{8.24} & \stackchange{38.60}{\uparrow}{2.20} & \stackchange{36.40}{\uparrow}{3.07} & \stackchange{\textbf{41.60}}{\uparrow}{4.20} \\
\midrule
\multirow{5}{*}{Env.}
& Base & 25.64 & \textbf{38.10} & 37.75 & \textbf{44.83} & 35.49 & 33.92 & 35.38 & 36.70 \\
 & + SFT & \stackchange{31.87}{\uparrow}{6.23} & \stackchange{37.64}{\downarrow}{0.46} & \stackchange{36.48}{\downarrow}{1.27} & \stackchange{44.64}{\downarrow}{0.19} & \stackchange{35.39}{\downarrow}{0.10} & \stackchange{32.31}{\downarrow}{1.61} & \stackchange{35.25}{\downarrow}{0.13} & \stackchange{36.23}{\downarrow}{0.47} \\
 & + GRPO & \stackchange{30.40}{\uparrow}{4.76} & \stackchange{35.60}{\downarrow}{2.50} & \stackchange{39.20}{\uparrow}{1.45} & \stackchange{41.13}{\downarrow}{3.70} & \stackchange{37.84}{\uparrow}{2.35} & \stackchange{38.60}{\uparrow}{4.68} & \stackchange{35.12}{\downarrow}{0.26} & \stackchange{37.87}{\uparrow}{1.17} \\
 & + OPSD & \stackchange{27.84}{\uparrow}{2.20} & \stackchange{37.87}{\downarrow}{0.23} & \stackchange{38.45}{\uparrow}{0.70} & \stackchange{43.47}{\downarrow}{1.36} & \stackchange{37.25}{\uparrow}{1.76} & \stackchange{38.74}{\uparrow}{4.82} & \stackchange{32.44}{\downarrow}{2.94} & \stackchange{37.40}{\uparrow}{0.70} \\
\rowcolor{opcadrow}
 & \textbf{+ OP-CAD} & \stackchange{\textbf{33.33}}{\uparrow}{7.69} & \stackchange{\textbf{38.10}}{\leftrightarrow}{0.00} & \stackchange{\textbf{41.47}}{\uparrow}{3.72} & \stackchange{44.44}{\downarrow}{0.39} & \stackchange{\textbf{39.22}}{\uparrow}{3.73} & \stackchange{\textbf{40.94}}{\uparrow}{7.02} & \stackchange{\textbf{38.19}}{\uparrow}{2.81} & \stackchange{\textbf{40.23}}{\uparrow}{3.53} \\
\midrule
\multirow{5}{*}{Speech}
& Base & 25.27 & 36.73 & 35.56 & 40.94 & 33.43 & 33.04 & 34.10 & 34.80 \\
 & + SFT & \stackchange{31.87}{\uparrow}{6.60} & \stackchange{38.10}{\uparrow}{1.37} & \stackchange{36.35}{\uparrow}{0.79} & \stackchange{42.69}{\uparrow}{1.75} & \stackchange{35.00}{\uparrow}{1.57} & \stackchange{34.65}{\uparrow}{1.61} & \stackchange{34.87}{\uparrow}{0.77} & \stackchange{36.20}{\uparrow}{1.40} \\
 & + GRPO & \stackchange{31.50}{\uparrow}{6.23} & \stackchange{36.05}{\downarrow}{0.68} & \stackchange{37.66}{\uparrow}{2.10} & \stackchange{39.57}{\downarrow}{1.37} & \stackchange{37.65}{\uparrow}{4.22} & \stackchange{37.43}{\uparrow}{4.39} & \stackchange{33.59}{\downarrow}{0.51} & \stackchange{36.87}{\uparrow}{2.07} \\
 & + OPSD & \stackchange{29.30}{\uparrow}{4.03} & \stackchange{36.51}{\downarrow}{0.22} & \stackchange{35.56}{\leftrightarrow}{0.00} & \stackchange{42.69}{\uparrow}{1.75} & \stackchange{34.71}{\uparrow}{1.28} & \stackchange{36.11}{\uparrow}{3.07} & \stackchange{29.89}{\downarrow}{4.21} & \stackchange{35.13}{\uparrow}{0.33} \\
\rowcolor{opcadrow}
 & \textbf{+ OP-CAD} & \stackchange{\textbf{32.97}}{\uparrow}{7.70} & \stackchange{\textbf{41.95}}{\uparrow}{5.22} & \stackchange{\textbf{39.06}}{\uparrow}{3.50} & \stackchange{\textbf{45.03}}{\uparrow}{4.09} & \stackchange{\textbf{38.73}}{\uparrow}{5.30} & \stackchange{\textbf{38.01}}{\uparrow}{4.97} & \stackchange{\textbf{36.02}}{\uparrow}{1.92} & \stackchange{\textbf{38.93}}{\uparrow}{4.13} \\
\bottomrule
\end{tabular}
}

\vspace{-6pt}
\end{table}

\subsection{Component Analysis}

\paragraph{Supervision signals.}
We compare audio removal sensitivity alone, corruption sensitivity alone, and their combination while fixing initialization, curriculum, and teacher targets (Table~\ref{tab:ablation-results}). Both individual signals improve all four corrupted-condition averages over uniform OPSD. The combined rule performs best in three aggregates; on OmniVideoBench competing speech, corruption sensitivity alone scores 39.00\% versus 38.93\% for OP-CAD. Both signals therefore provide useful allocation criteria, and their combination improves most evaluated noise-family averages.

\begin{table}[!htbp]
\caption{Accuracy (\%) with single-signal and combined allocation. Bold marks the best result.}
\label{tab:ablation-results}
\centering
{\small
\setlength{\tabcolsep}{0.60pt}
\renewcommand{\arraystretch}{1.12}
\begin{tabularx}{\columnwidth}{l*{6}{>{\raggedleft\arraybackslash}X}}
\toprule
& \multicolumn{3}{c}{Daily-Omni} & \multicolumn{3}{c}{OmniVideoBench} \\
\cmidrule(lr){2-4}\cmidrule(lr){5-7}
Variant & Clean & Env. & Speech & Clean & Env. & Speech \\
\midrule
OPSD & 69.34 & 63.49 & 60.62 & 41.40 & 37.40 & 35.13 \\
Audio-rem. proxy & 69.34 & 64.47 & 61.71 & 40.70 & 39.63 & 38.50 \\
Corr.-sens. proxy & 68.76 & 65.41 & 62.13 & \textbf{43.80} & 39.73 & \textbf{39.00} \\
\midrule
\rowcolor{opcadrow}
\textbf{OP-CAD} & \textbf{70.09} & \textbf{66.97} & \textbf{63.55} & 41.60 & \textbf{40.23} & 38.93 \\
\bottomrule
\end{tabularx}
}
\vspace{-6pt}
\end{table}

\paragraph{Noise curriculum.}
\label{sec:curriculum-ablation}
We compare OP-CAD's 10, 5, then 0 dB curriculum with randomly mixed SNRs under matched total training steps and data budgets (Table~\ref{tab:curriculum-order}). Curriculum learning improves accuracy under both interference types on both benchmarks. From the displayed scores, environmental noise and competing speech gains are 1.81 and 0.88 percentage points on Daily-Omni, and 1.96 and 1.80 points on OmniVideoBench. Clean accuracy is higher with the curriculum on Daily-Omni but lower on OmniVideoBench. With the training budget held fixed, the noisy-condition gains support the mild-to-severe curriculum for improving acoustic robustness.

\begin{table}[!htbp]
\caption{Curriculum order ablation: accuracy (\%) under matched training budgets.}
\label{tab:curriculum-order}
\centering
{\small
\setlength{\tabcolsep}{4pt}
\renewcommand{\arraystretch}{1.18}
\begin{tabularx}{\linewidth}{l*{6}{>{\raggedleft\arraybackslash}X}}
\toprule
\multirow{2}{*}{Training schedule} & \multicolumn{3}{c}{Daily-Omni} & \multicolumn{3}{c}{OmniVideoBench} \\
\cmidrule(lr){2-4}\cmidrule(lr){5-7}
& Clean & Env. & Speech & Clean & Env. & Speech \\
\midrule
Randomly mixed & 69.93 & 65.16 & 62.67 & \textbf{42.30} & 38.27 & 37.13 \\
\rowcolor{opcadrow}
\textbf{Curriculum learning} & \textbf{70.09} & \textbf{66.97} & \textbf{63.55} & 41.60 & \textbf{40.23} & \textbf{38.93} \\
\bottomrule
\end{tabularx}
}
\vspace{-6pt}
\end{table}

\subsection{Prediction Stability under Interference}

\begin{figure}[!htbp]
    \vspace{-6pt}
    \centering
    \includegraphics[width=0.90\linewidth]{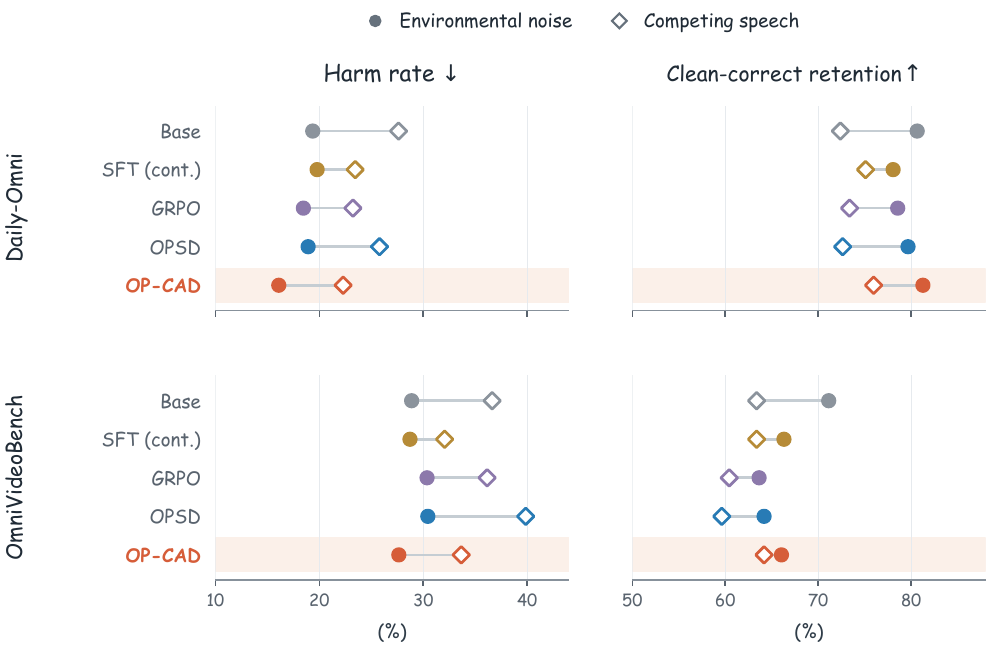}
    \vspace{-4pt}
    \caption{Prediction stability at 0 dB: harm rate (left) and clean-correct retention (right).}
    \label{fig:paired-transitions}
\vspace{-8pt}
\end{figure}

Aggregate noisy accuracy combines retained clean successes and newly correct answers. Figure~\ref{fig:paired-transitions} separates these outcomes using two measures at 0 dB. Harm rate measures the fraction of each model's own clean-correct answers that become incorrect; clean-correct retention measures noisy accuracy on the fixed set of questions that Base answers correctly under clean audio. The former assesses each model's vulnerability to interference, while the latter evaluates all models on the same questions that Base answers correctly under clean audio.

OP-CAD has lower harm rates than OPSD in all four conditions and the lowest rate among the compared methods in three. The largest reduction occurs on OmniVideoBench competing speech, from 39.86\% to 33.65\%, with a paired difference of $-6.20$ points. SFT has a slightly lower harm rate of 32.05\% in this condition.

OP-CAD achieves the highest retention on Base's clean-correct questions in three conditions. Under OmniVideoBench environmental noise, Base retains 71.12\% accuracy versus 66.04\% for OP-CAD, indicating that improved full test accuracy does not entail better preservation of every baseline success.

The overall degradation is also smaller: under competing speech at 0 dB on OmniVideoBench, the net loss of correct answers from clean to noisy input is 34 for OP-CAD versus 71 for OPSD. Together with the clean results in Table~\ref{tab:main-results}, this shows improved resistance to interference without an observed decrease in aggregate clean accuracy.

\subsection{Training Example of Supervision Allocation}

\begin{figure}[!htb]
    \vspace{-6pt}
    \centering
    \includegraphics[width=0.92\linewidth]{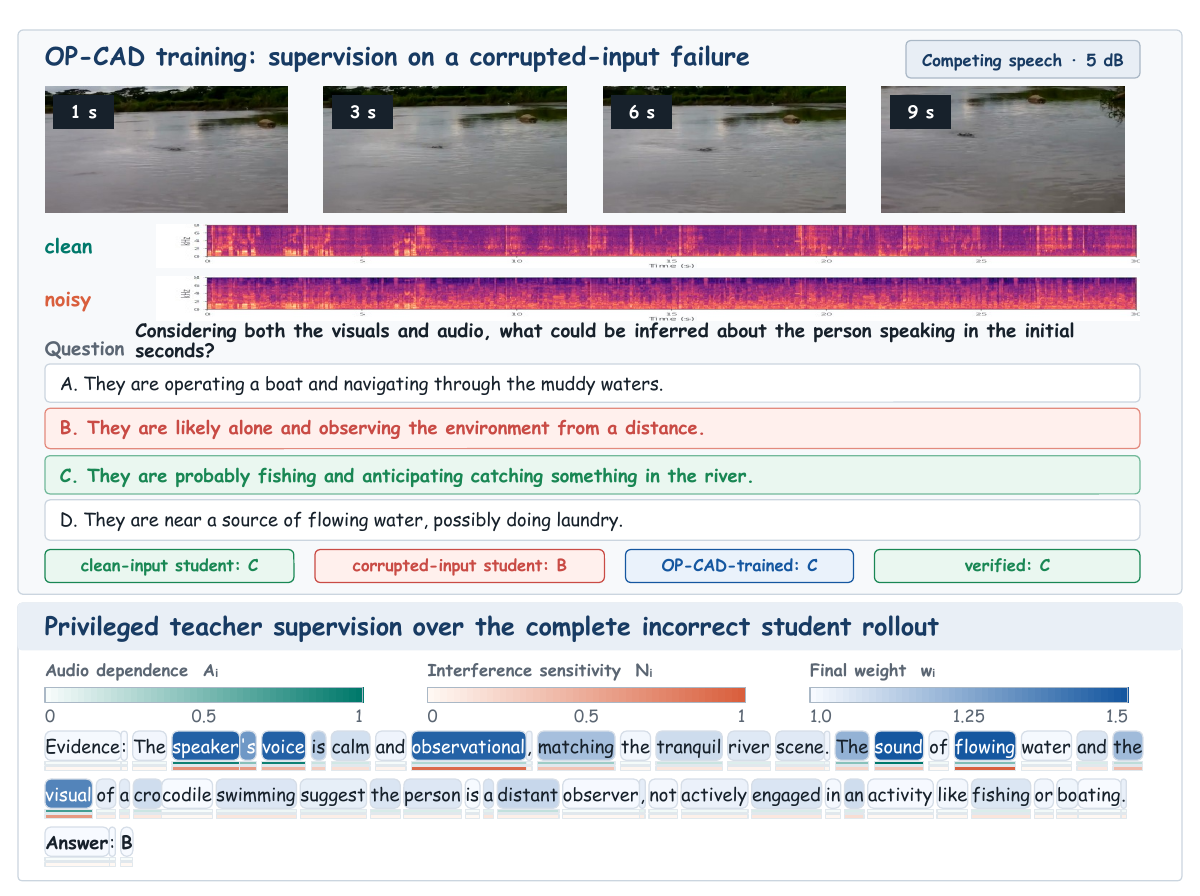}
    \vspace{-4pt}
    \caption{Token-level supervision allocation on an OP-CAD training rollout.}
    \label{fig:token-allocation}
\vspace{-8pt}
\end{figure}

Figure~\ref{fig:token-allocation} shows how the two signals distribute supervision on a recorded training rollout under competing speech at 5 dB. Audio removal sensitivity is strongest around ``voice'' and ``sound,'' whereas corruption sensitivity peaks around ``observational'' and ``flowing.'' The signals therefore emphasize different positions in this example. Their combination retains both sets of peaks while preserving baseline supervision elsewhere, illustrating how the weighting rule operates along a student's response under shared visual context.

\FloatBarrier

\Needspace{8\baselineskip}
\section{Conclusion}

OP-CAD weights clean-teacher supervision by the teacher's sensitivity to audio removal and corruption. On two audio-visual benchmarks, it improves accuracy on noisy input over uniform OPSD under matched training conditions, with no observed decrease in aggregate clean accuracy. Single-signal ablations show that the two comparisons are useful but that their combination is not best in every condition. The method changes supervision during training; inference uses the student alone, without a clean audio reference, verified answer, or additional teacher passes. The current evidence is limited to one backbone, one run per method, and controlled acoustic corruptions.

\newpage
\bibliography{references}
\bibliographystyle{plainnat}
\clearpage
\appendix
\raggedbottom
\renewcommand{\topfraction}{0.9}
\renewcommand{\textfraction}{0.08}
\renewcommand{\floatpagefraction}{0.85}
\setlength{\textfloatsep}{16pt plus 2pt minus 2pt}
\setlength{\floatsep}{14pt plus 2pt minus 2pt}
\setlength{\intextsep}{14pt plus 2pt minus 2pt}
\makeatletter
\setlength{\@fptop}{0pt}
\setlength{\@fpsep}{14pt}
\setlength{\@fpbot}{0pt plus 1fil}
\makeatother

The appendix gives implementation details (Section~\ref{app:implementation}), data construction (Section~\ref{app:data}), additional results and diagnostics (Section~\ref{app:results}), and reproduction settings (Section~\ref{app:reproducibility}).

\FloatBarrier
\section{Implementation Details}
\label{app:implementation}

\Needspace{5\baselineskip}
\subsection{Full-Vocabulary Distributions and Response Handling}

All teacher and student distributions in Section~\ref{sec:method} use the same complete vocabulary $\mathcal V$. For logits $\ell$, the temperature distribution is
\begin{equation}
p_\tau(z;\ell)=\frac{\exp(\ell_z/\tau)}{\sum_{z'\in\mathcal V}\exp(\ell_{z'}/\tau)},
\qquad z\in\mathcal V.
\end{equation}
The reverse-KL target loss and both allocation JS divergences sum over $\mathcal V$. Teacher logits are detached; gradients flow only through the student. Log-softmax and log-space mixture calculations provide stable normalization across the complete vocabulary.

The corrupted-input student draws one current-policy response of at most 128
new tokens. The prompt requests an \texttt{Evidence:} field followed by an
\texttt{Answer:} field whose valid option letters are determined by the
example. All teacher contexts score this same trajectory. Every
generated position contributes to the objective, including positions in an
incomplete or truncated response; such responses are not regenerated. Evidence
presence, answer validity, response length, and truncation are logged only for
monitoring.

\Needspace{5\baselineskip}
\subsection{Numerical and Reduction Details}

For either score $Z\in\{A,N\}$, we use $\widetilde Z_i=Z_i/\max_j Z_j$ when the response maximum exceeds $10^{-12}$, and zero otherwise. This threshold prevents division by numerical zero. The loss sums weighted token losses across the distributed mini-batch and divides by the total weight. The paper configuration uses one microbatch per optimizer update. Sampled prefixes, teacher distributions, and weights are detached; gradients flow only through the student distribution.

\FloatBarrier
\section{Paired Robustness Dataset}
\label{app:data}

\Needspace{5\baselineskip}
\subsection{Dataset Construction and Supervision}

We use WorldSense, OmniBench, and the AVQA subset of OmniInstruct-v1 for training, with the screening rule in Section~\ref{sec:data-routing}. Daily-Omni and OmniVideoBench are used only for evaluation. The local robustness partition contains 4,660 corrupted acoustic views of 1,809 distinct source questions. Table~\ref{tab:training-source-counts} gives its source and SNR composition. Applying the warm-up routing criterion to the archived screened pool at the same three SNRs yields 1,790 views of 615 questions (Table~\ref{tab:warmup-source-counts}). A question may contribute several corrupted views, so view counts and question counts are reported separately.

Both SFT stages use Gemini 3 Flash-generated evidence \citep{gemini3flash} paired with the dataset's verified answer. The evidence generator is separate from the frozen Qwen teacher used for distillation; Appendix~\ref{app:gemini-evidence} specifies its inputs and a reproduction template. All generated warm-up evidence underwent human review for support from the audio-visual evidence. This review checks the explanation rather than relying on answer-label agreement alone.

\begin{table}[!htbp]
\centering
\caption{Robustness-data partition after paired screening. SNR columns count corrupted acoustic views.}
\label{tab:training-source-counts}
\normalsize
\setlength{\tabcolsep}{9pt}
\begin{tabular}{lrrrrr}
\toprule
Source & Questions & 10 dB & 5 dB & 0 dB & Total views \\
\midrule
WorldSense & 619 & 416 & 524 & 712 & 1,652 \\
OmniBench & 291 & 241 & 281 & 317 & 839 \\
OmniInstruct-v1 (AVQA) & 899 & 587 & 697 & 885 & 2,169 \\
\midrule
Total & 1,809 & 1,244 & 1,502 & 1,914 & 4,660 \\
\bottomrule
\end{tabular}
\end{table}

\begin{table}[!htbp]
\centering
\caption{Warm-up data after paired screening.}
\label{tab:warmup-source-counts}
\normalsize
\setlength{\tabcolsep}{9pt}
\begin{tabular}{lrrrrr}
\toprule
Source & Questions & 10 dB & 5 dB & 0 dB & Total views \\
\midrule
WorldSense & 276 & 210 & 205 & 438 & 853 \\
OmniBench & 97 & 61 & 65 & 134 & 260 \\
OmniInstruct-v1 (AVQA) & 242 & 160 & 167 & 350 & 677 \\
\midrule
Total & 615 & 431 & 437 & 922 & 1,790 \\
\bottomrule
\end{tabular}
\end{table}

The main experiments use environmental and competing-speech corruption at
$\{10,5,0\}$ dB. For a normalized clean waveform, the procedure applies a
measured room impulse response, samples the reverberant wet ratio from
$[0.30,0.80]$, simulates an 8-kHz $\mu$-law channel followed by resampling, and
RMS-scales the additive source to the target SNR. \par
\begin{samepage}
Training uses 61 measured RIRs; evaluation uses 11 disjoint measured RIRs.\par
\end{samepage}
 The environmental pool contains
80,000 training and 20,000 evaluation mixtures. The competing-speech indices
contain 16,922 training and 618 evaluation mixtures with disjoint VoxCeleb2
speakers and utterance files. The video, question, option order, answer, and
temporal sampling remain fixed within each clean--corrupted pair.

\FloatBarrier
\section{Complete Robustness Results}
\label{app:results}

Tables~\ref{tab:supp-base-breakdown} and~\ref{tab:supp-per-snr-accuracy} report Base performance by benchmark partition and all compared methods at each SNR. OP-CAD exceeds OPSD in all 12 benchmark--interference--SNR conditions.

\begin{table}[!htbp]
\caption{Base accuracy (\%) by benchmark category and duration under paired acoustic conditions.}
\label{tab:supp-base-breakdown}
\label{tab:noise-accuracy-trend}
\centering
\small
\setlength{\tabcolsep}{2.5pt}
\begin{adjustbox}{max width=\linewidth}
\begin{tabular}{@{}lrrrrrrrrr@{}}
\toprule
\multicolumn{10}{c}{\textit{Daily-Omni}} \\
Condition & AV Align & Comp. & Ctx. Und. & Evt. Seq. & Infer. & Reas. & 30s & 60s & Avg. \\
\midrule
Clean     & 58.82 & 77.86 & 60.62 & 60.13 & 81.82 & 81.14 & 68.01 & 67.45 & 67.75 \\
Env. 10   & 52.94 & 70.99 & 60.62 & 55.88 & 79.87 & 74.29 & 65.53 & 61.09 & 63.49 \\
Env. 5    & 47.48 & 72.52 & 59.59 & 54.25 & 76.62 & 70.86 & 61.98 & 60.00 & 61.07 \\
Env. 0    & 47.48 & 69.47 & 59.59 & 53.92 & 71.43 & 68.00 & 59.51 & 59.64 & 59.57 \\
Speech 10 & 52.52 & 70.23 & 59.07 & 52.94 & 75.32 & 72.57 & 61.82 & 61.09 & 61.49 \\
Speech 5  & 46.22 & 65.65 & 59.07 & 50.98 & 72.08 & 65.14 & 58.11 & 57.27 & 57.73 \\
Speech 0  & 43.28 & 64.89 & 55.96 & 46.73 & 66.88 & 61.71 & 54.71 & 53.82 & 54.30 \\
\bottomrule
\end{tabular}
\end{adjustbox}

\setlength{\tabcolsep}{3.5pt}
\begin{adjustbox}{max width=\linewidth}
\begin{tabular}{@{}lrrrrrrrr@{}}
\toprule
\multicolumn{9}{c}{\textit{OmniVideoBench}} \\
\multirow{2}{*}{Condition} & \multicolumn{3}{c}{Audio Type} & \multicolumn{4}{c}{Video Duration} & \multirow{2}{*}{Avg.} \\
\cmidrule(lr){2-4}\cmidrule(lr){5-8}
& Music & Sound & Speech & (0,1] min & (1,5] min & (5,10] min & (10,30] min & \\
\midrule
Clean     & 30.77 & 32.65 & 39.11 & 49.12 & 35.29 & 36.40 & 33.33 & 37.40 \\
Env. 10   & 26.37 & 38.10 & 38.19 & 46.20 & 34.41 & 35.53 & 36.02 & 37.10 \\
Env. 5    & 25.27 & 38.78 & 37.40 & 42.69 & 36.47 & 32.02 & 36.40 & 36.50 \\
Env. 0    & 25.27 & 37.41 & 37.66 & 45.61 & 35.59 & 34.21 & 33.72 & 36.50 \\
Speech 10 & 26.37 & 38.10 & 37.66 & 42.69 & 34.71 & 37.28 & 34.87 & 36.70 \\
Speech 5  & 23.08 & 35.37 & 35.96 & 39.77 & 33.53 & 31.14 & 36.02 & 34.70 \\
Speech 0  & 26.37 & 36.73 & 33.07 & 40.35 & 32.06 & 30.70 & 31.42 & 33.00 \\
\bottomrule
\end{tabular}
\end{adjustbox}
\end{table}

\begin{table}[!htbp]
\caption{Full-test accuracy (\%) at each noise severity.}
\label{tab:supp-per-snr-accuracy}
\centering
\small
\setlength{\tabcolsep}{4pt}
\begin{adjustbox}{max width=\linewidth}
\begin{tabular}{llrrrrrrr}
\toprule
Benchmark & Method & Clean & Env. 10 & Env. 5 & Env. 0 & Speech 10 & Speech 5 & Speech 0 \\
\midrule
\multirow{5}{*}{Daily-Omni}
& Base & 67.75 & 63.49 & 61.07 & 59.57 & 61.49 & 57.73 & 54.30 \\
& SFT & 68.42 & 63.41 & 62.74 & 59.48 & 62.24 & 57.89 & 57.23 \\
& GRPO & 68.34 & 64.41 & 64.41 & 59.90 & 63.49 & 58.56 & 57.73 \\
& OPSD & 69.34 & 65.83 & 63.66 & 60.99 & 64.16 & 61.57 & 56.14 \\
\rowcolor{opcadrow}
& OP-CAD & 70.09 & 69.42 & 67.08 & 64.41 & 67.08 & 63.74 & 59.82 \\
\midrule
\multirow{5}{*}{OmniVideoBench}
& Base & 37.40 & 37.10 & 36.50 & 36.50 & 36.70 & 34.70 & 33.00 \\
& SFT & 39.00 & 35.30 & 36.90 & 36.50 & 37.00 & 35.00 & 36.60 \\
& GRPO & 41.50 & 38.70 & 37.80 & 37.10 & 38.80 & 36.10 & 35.70 \\
& OPSD & 41.40 & 37.40 & 36.20 & 38.60 & 36.50 & 34.60 & 34.30 \\
\rowcolor{opcadrow}
& OP-CAD & 41.60 & 40.10 & 39.40 & 41.20 & 39.10 & 39.50 & 38.20 \\
\bottomrule
\end{tabular}
\end{adjustbox}
\end{table}

\Needspace{5\baselineskip}
\subsection{Robustness across Categories and Durations}

The category and duration accuracies are reported in Tables~\ref{tab:daily-category-results} and~\ref{tab:omnivideo-category-results}. Here we provide the partition sizes. Each corrupted-condition accuracy averages the 10, 5, and 0 dB views within its interference family.

\begin{table}[!htbp]
\caption{Test-set counts by category and duration; these are separate partitions of the same examples.}
\label{tab:supp-category-counts}
\centering
\small
\setlength{\tabcolsep}{4.5pt}
\begin{adjustbox}{max width=\linewidth}
\begin{tabular}{@{}p{0.20\linewidth}p{0.15\linewidth}p{0.57\linewidth}@{}}
\toprule
Benchmark & Partition & Group counts \\
\midrule
Daily-Omni & Question type &
AV Align 238; Comp. 131; Ctx. Und. 193; Evt. Seq. 306; Infer. 154; Reas. 175 \\
Daily-Omni & Duration & 30s 647; 60s 550 \\
OmniVideoBench & Audio type & Music 91; Sound 147; Speech 762 \\
OmniVideoBench & Duration &
(0,1] min 171; (1,5] min 340; (5,10] min 228; (10,30] min 261 \\
\bottomrule
\end{tabular}
\end{adjustbox}
\end{table}

Official metadata assigns five OmniVideoBench records slightly longer than 30 minutes to the final published duration bin; we retain this assignment.

\Needspace{5\baselineskip}
\subsection{Preserving Clean-Correct Answers}

\paragraph{Paired metrics.}
Let $B_D$ be the questions in benchmark $D$ that Base answers correctly under clean audio, and let $y_i$ be the correct option for question $i$. The clean-correct retention of model $M$
under condition $c$ is
\begin{equation}
    R(M,c)=\frac{1}{|B_D|}\sum_{i\in B_D}
    \mathbf{1}[M(i,c)=y_i].
\end{equation}
This fixed denominator measures preservation of examples solved by the base
model and must be read alongside complete-test accuracy.

For each model, we also count the transitions between clean and corrupted correctness. Write $N_{uv}$ for the number of questions with clean correctness $u$ and corrupted correctness $v$, where 1 means correct. Thus $N_{11}$ stays correct, $N_{10}$ becomes wrong, $N_{01}$ becomes correct, and $N_{00}$ stays wrong. Harm rate $H$, recovery rate $G$, and net accuracy change are
\begin{equation}
\begin{aligned}
    H&=\frac{N_{10}}{N_{11}+N_{10}}, &
    G&=\frac{N_{01}}{N_{01}+N_{00}},\\
    \Delta\mathrm{Acc}&=100\frac{N_{01}-N_{10}}{|D|}.
\end{aligned}
\end{equation}
Here $H$ is the harm rate among clean-correct examples, $G$ is the recovery
rate among clean-wrong examples, and $\Delta\mathrm{Acc}$ is the paired
accuracy change in percentage points. A nonzero $N_{01}$ does not by itself
imply that corruption is beneficial; it must be interpreted with $N_{10}$ and
the net change.

Tables~\ref{tab:supp-daily-per-snr} and~\ref{tab:supp-transitions} report retention at each SNR and correctness transitions at 0 dB. OP-CAD exceeds OPSD
in every displayed clean-correct retention column on both benchmarks.

\begin{table}[!htbp]
\caption{Clean-correct retention (\%) on Base's fixed clean-correct set at each SNR: Daily-Omni (811 questions, top) and OmniVideoBench (374 questions, bottom).}
\label{tab:supp-daily-per-snr}
\centering
\normalsize
\setlength{\tabcolsep}{6pt}
\begin{adjustbox}{max width=\linewidth}
\begin{tabular}{lrrrrrrr}
\toprule
 Method & Clean & Env 10 & Env 5 & Env 0 & Speech 10 & Speech 5 & Speech 0 \\
\midrule
 Base & 100.00 & 87.67 & 84.46 & 80.64 & 84.59 & 78.55 & 72.38 \\
 SFT & 90.14 & 84.71 & 83.60 & 78.05 & 83.85 & 77.68 & 75.09 \\
 OPSD & 88.78 & 85.45 & 82.37 & 79.65 & 83.48 & 80.52 & 72.63 \\
\rowcolor{opcadrow}
 OP-CAD & 89.40 & 88.53 & 85.45 & 81.26 & 85.20 & 81.50 & 75.96 \\
\bottomrule
\end{tabular}
\end{adjustbox}
\par\vspace{8pt}
\label{tab:supp-ovb-per-snr}
\centering
\normalsize
\setlength{\tabcolsep}{6pt}
\begin{adjustbox}{max width=\linewidth}
\begin{tabular}{lrrrrrrr}
\toprule
 Method & Clean & Env 10 & Env 5 & Env 0 & Speech 10 & Speech 5 & Speech 0 \\
\midrule
 Base & 100.00 & 73.80 & 71.12 & 71.12 & 71.93 & 66.31 & 63.37 \\
 SFT & 74.60 & 66.31 & 67.65 & 66.31 & 66.04 & 63.10 & 63.37 \\
 OPSD & 72.99 & 64.44 & 64.71 & 64.17 & 63.10 & 61.23 & 59.63 \\
\rowcolor{opcadrow}
 OP-CAD & 74.33 & 67.11 & 67.91 & 66.04 & 65.78 & 67.11 & 64.17 \\
\bottomrule
\end{tabular}
\end{adjustbox}
\end{table}

\begin{table}[!htbp]
\caption{Correctness transitions at 0 dB. $H/G$: harm/recovery rates (\%); $\Delta$: clean-to-noisy accuracy change (percentage points).}
\label{tab:supp-transitions}
\centering
\normalsize
\setlength{\tabcolsep}{8pt}
\begin{adjustbox}{max width=\linewidth}
\begin{tabular}{llrrrrrrr}
\toprule
\multicolumn{9}{c}{\textit{Daily-Omni}} \\
 Method & Cond. & $N_{11}$ & $N_{10}$ & $N_{01}$ & $N_{00}$ & $H$ & $G$ & $\Delta$ \\
\midrule
 Base & Env. 0 & 654 & 157 & 59 & 327 & 19.36 & 15.28 & -8.19 \\
 Base & Speech 0 & 587 & 224 & 63 & 323 & 27.62 & 16.32 & -13.45 \\
 SFT & Env. 0 & 657 & 162 & 55 & 323 & 19.78 & 14.55 & -8.94 \\
 SFT & Speech 0 & 627 & 192 & 58 & 320 & 23.44 & 15.34 & -11.19 \\
 OPSD & Env. 0 & 673 & 157 & 57 & 310 & 18.92 & 15.53 & -8.35 \\
 OPSD & Speech 0 & 616 & 214 & 56 & 311 & 25.78 & 15.26 & -13.20 \\
\rowcolor{opcadrow}
 OP-CAD & Env. 0 & 704 & 135 & 67 & 291 & 16.09 & 18.72 & -5.68 \\
\rowcolor{opcadrow}
 OP-CAD & Speech 0 & 652 & 187 & 64 & 294 & 22.29 & 17.88 & -10.28 \\
\bottomrule
\end{tabular}
\end{adjustbox}
\par\medskip
\begin{adjustbox}{max width=\linewidth}
\begin{tabular}{llrrrrrrr}
\toprule
\multicolumn{9}{c}{\textit{OmniVideoBench}} \\
 Method & Cond. & $N_{11}$ & $N_{10}$ & $N_{01}$ & $N_{00}$ & $H$ & $G$ & $\Delta$ \\
\midrule
 Base & Env. 0 & 266 & 108 & 99 & 527 & 28.88 & 15.81 & -0.90 \\
 Base & Speech 0 & 237 & 137 & 93 & 533 & 36.63 & 14.86 & -4.40 \\
 SFT & Env. 0 & 278 & 112 & 87 & 523 & 28.72 & 14.26 & -2.50 \\
 SFT & Speech 0 & 265 & 125 & 101 & 509 & 32.05 & 16.56 & -2.40 \\
 OPSD & Env. 0 & 288 & 126 & 98 & 488 & 30.43 & 16.72 & -2.80 \\
 OPSD & Speech 0 & 249 & 165 & 94 & 492 & 39.86 & 16.04 & -7.10 \\
\rowcolor{opcadrow}
 OP-CAD & Env. 0 & 301 & 115 & 111 & 473 & 27.64 & 19.01 & -0.40 \\
\rowcolor{opcadrow}
 OP-CAD & Speech 0 & 276 & 140 & 106 & 478 & 33.65 & 18.15 & -3.40 \\
\bottomrule
\end{tabular}
\end{adjustbox}
\end{table}

\begin{samepage}
At 0 dB, OP-CAD reduces the net clean-to-noisy accuracy loss relative to OPSD: from 8.35 to 5.68 points and from 13.20 to 10.28 points on Daily-Omni, and from 2.80 to 0.40 points and from 7.10 to 3.40 points on OmniVideoBench, for environmental noise and competing speech respectively. The transition counts also show why a small net loss can coexist with many previously correct answers becoming wrong.\par
\end{samepage}

\Needspace{5\baselineskip}
\subsection{Context Sensitivity}

We scored 639 saved tokens from eight auxiliary diagnostic trajectories under seven contexts using the frozen SFT warm-up teacher used in OP-CAD training. Table~\ref{tab:supp-context-forward} reports full-vocabulary disagreement after removing audio, corrupting it, or removing the answer. These fixed-response comparisons measure the teacher's context sensitivity on the sampled prefixes. The answer-free contrasts match the allocation contexts in Section~\ref{sec:method}; contrasts containing the answer additionally examine privileged conditioning. This small diagnostic sample does not establish how frequently these effects occur across the training set.

\begin{table}[!htbp]
\caption{Teacher predictive disagreement. Flip: top-1 disagreement rate (\%).}
\label{tab:supp-context-forward}
\centering
\normalsize
\setlength{\tabcolsep}{3pt}
\begin{adjustbox}{max width=\linewidth}
\begin{tabular}{@{}>{\raggedright\arraybackslash}p{0.45\linewidth}rrrr@{}}
\toprule
Reference $\rightarrow$ comparison
& \multicolumn{2}{c}{All} & \multicolumn{2}{c}{Evidence} \\
\cmidrule(lr){2-3}\cmidrule(lr){4-5}
& JS & Flip & JS & Flip \\
\midrule
Clean+answer $\rightarrow$ visual+answer & 0.0527 & 17.68 & 0.0592 & 19.40 \\
Clean+answer $\rightarrow$ corrupted+answer & 0.0201 & 11.42 & 0.0221 & 12.46 \\
Clean+answer $\rightarrow$ text+answer & 0.1046 & 27.70 & 0.1165 & 30.07 \\
Clean+answer $\rightarrow$ clean, no answer & 0.0113 & 8.29 & 0.0092 & 8.19 \\
Clean, no answer $\rightarrow$ corrupted, no answer & 0.0233 & 11.11 & 0.0256 & 12.10 \\
Clean, no answer $\rightarrow$ visual, no answer & 0.0540 & 17.06 & 0.0609 & 19.04 \\
\bottomrule
\end{tabular}
\end{adjustbox}
\end{table}

\Needspace{9\baselineskip}
\subsection{Training Efficiency}

Table~\ref{tab:supp-training-cost} gives the cost of the robustness-training runs, excluding shared SFT warm-up, offline evidence generation, and evaluation. OP-CAD uses one rollout per example and four teacher contexts. It takes 36.5\% more GPU-hours than OPSD and about 65\% fewer than the eight-rollout GRPO configuration. These are measured costs of the listed runs, not comparisons at equal compute.

\begin{table}[!htbp]
\caption{Measured cost of robustness training.
$^\ast$GRPO has no privileged teacher but does perform its reference-policy KL
forward. GPU-hours assume all 32 configured GPUs remain allocated over the
stated wall-time basis.}
\label{tab:supp-training-cost}
\centering
\normalsize
\setlength{\tabcolsep}{10pt}
\begin{adjustbox}{max width=\linewidth}
\begin{tabular}{lrrr}
\toprule
 Method & Wall h & GPU-h & Peak GiB/rank \\
\midrule
 OPSD & 12.29 & 393.14 & 115.3 \\
\rowcolor{opcadrow}
 OP-CAD & 16.77 & 536.50 & 116.05 \\
 GRPO$^\ast$ & 48.38 & 1,548.05 & 123.8 \\
\bottomrule
\end{tabular}
\end{adjustbox}
\end{table}

\FloatBarrier
\Needspace{6\baselineskip}
\section{Reproducibility Details}
\label{app:reproducibility}

\Needspace{5\baselineskip}
\subsection{Artifact Availability and Computing Environment}

The accompanying anonymous source package provides the training and evaluation implementation and configuration templates. The acoustic corruption procedure and parameters are described in Appendix~\ref{app:data}.

Fine-tuning used 32 NVIDIA L20X GPUs (143,166 MiB each) across four Linux nodes with AMD EPYC 9T24 CPUs. The stack used
Python 3.11, CUDA 12, PyTorch 2.9.1, Accelerate 1.13.0, DeepSpeed 0.18.9,
Datasets 3.6.0, PEFT 0.14.0, TRL 0.29.1, Tokenizers 0.21.4, and FlashAttention
2.8.3.

\begin{table}[!htbp]
\caption{Main training configuration.}
\label{tab:supp-training-config}
\centering
\footnotesize
\setlength{\tabcolsep}{3pt}
\begin{tabular*}{\textwidth}{@{\extracolsep{\fill}}>{\raggedright\arraybackslash}p{0.18\textwidth}>{\raggedright\arraybackslash}p{0.27\textwidth}>{\raggedright\arraybackslash}p{0.18\textwidth}>{\raggedright\arraybackslash}p{0.27\textwidth}@{}}
\toprule
Setting & Value & Setting & Value \\
\midrule
Backbone & Qwen3-Omni-Instruct
& Model / data-order seed & 42 / 42 \\
Student initialization & Shared SFT warm-up checkpoint (1 epoch)
& Gradient accumulation & 1 \\
Teacher & Frozen copy of the SFT warm-up checkpoint
& Video sampling & 1 FPS, at most 256 frames \\
Trainable adaptation & LoRA on the language backbone and multimodal projectors
& Per-frame pixel budget & 262,144 \\
Frozen components & Acoustic and visual encoders
& Audio-duration cap & None \\
LoRA rank / scale & 64 / 128
& Training objective & Counterfactual-weighted reverse-KL OPSD \\
LoRA dropout & 0.05
& Shared distribution temperature & 1 \\
Learning rate & $10^{-5}$
& Allocation gain / maximum weight & $\lambda=0.5$ / $w_{\max}=1.5$ \\
LR warmup ratio & 0.03
& Vocabulary domain & Complete vocabulary for loss and allocation \\
Optimizer / precision & Fused AdamW / bfloat16
& Generation limit & 128 new tokens \\
Effective global batch & 32
& Response handling & One response; no filtering or resampling \\
Curriculum & 10 $\rightarrow$ 5 $\rightarrow$ 0 dB
& Batch construction & Length-aware multimodal batching \\
Distributed training & 4 nodes, 32 GPUs, ZeRO-2
& Epochs per SNR stage & 1 / 1 / 1 (10 / 5 / 0 dB) \\
\bottomrule
\end{tabular*}
\end{table}

\Needspace{5\baselineskip}
\FloatBarrier
\Needspace{20\baselineskip}
\subsection{Prompt and Scoring Protocol}

The student receives video, degraded audio, and the following text, with allowed letters set by the option count:
\begin{quote}
\raggedright
\texttt{You are given a video. Based on the content of the video, answer the
following question:}\\
\texttt{Question: \{question\}}\\
\texttt{Options: \{labeled options\}}\\
\texttt{Respond in this exact format:}\\
\texttt{Evidence: audio-visual evidence explaining why this answer option is
correct, no more than 100 English words}\\
\texttt{Answer: <one option letter>}\\
\texttt{Replace <one option letter> with exactly one option letter from
\{allowed letters\}. Do not output ``X''. Do not output a long reasoning
process.}
\end{quote}

The privileged clean teacher inserts the following private block before the
response-format instruction:
\begin{quote}
\raggedright
\texttt{Private supervision (do not quote or mention it in the response):}\\
\texttt{<correct\_option>\{answer\}}\\
\texttt{</correct\_option>}\\
\texttt{Use the private option only to verify the evidence and final choice.}
\end{quote}
The teacher is also instructed not to mention private supervision or input audio quality. Allocation contexts share the student text and omit the private block. Both SFT stages pair the student prompt with Gemini-generated evidence and the verified answer.

Evaluation uses the same response structure. The released evaluator samples Daily-Omni at 1 FPS and OmniVideoBench at 2 FPS, with at most 256 frames in both cases. We extract the designated answer field, with conservative fallback extraction; isolated letters in the evidence do not count as answers. Unparseable outputs count as incorrect and are not regenerated. Deterministic generation uses a 512-token limit and an 8-gram repetition constraint on newly generated tokens.

\Needspace{8\baselineskip}
\subsection{Offline Evidence Generation}
\label{app:gemini-evidence}

Gemini 3 Flash is called through its API with the clean audio, the paired video sampled at 1 FPS (matching training), the question, all labeled options, and the dataset's verified option letter. This offline generator is distinct from the frozen Qwen teacher used during distillation. The generated evidence is paired with the verified answer for supervised training.

\paragraph{Generation settings.}
Generation uses temperature 0.8 and a maximum output length of 512 tokens. The following prompt template specifies the inputs and output format for reproduction.

\paragraph{System instruction.}
\begin{quote}\raggedright
Describe concise, observable evidence for audio-visual question answering. Ground every claim in the supplied audio and video; never invent observations to match the reference answer, including inaudible speech, unseen actions, or unsupported identities.
\end{quote}

\Needspace{14\baselineskip}
\paragraph{User prompt.}
\begin{quote}\raggedright
\texttt{Question: \{question\}}\\
\texttt{Options: \{labeled options\}}\\
\texttt{Reference answer: \{verified option letter\}}\\
Inspect the supplied clean audio and video. Identify the observable acoustic and visual cues supporting the reference answer. Use at most 100 English words and do not mention access to a reference answer. Return exactly:\\
\texttt{Evidence: \{observable evidence\}}\\
\texttt{Answer: \{verified option letter\}}\\
\end{quote}

\paragraph{Export validation and human review.}
The export routine checks the Evidence/Answer structure, agreement with the dataset answer, nonempty evidence, a 100-word limit, and a list of invalid generic phrases. All generated warm-up evidence underwent human review to check whether the audio-visual evidence supported the explanation.

\Needspace{5\baselineskip}
\subsection{Configuration and Checkpoint Selection}

We evaluate final curriculum checkpoints using the initialization and stages in Section~\ref{sec:warmup-curriculum}; neither benchmark is used for hyperparameter tuning or checkpoint selection.

Supervised warm-up trains for one epoch. Each SNR stage (10, 5, and 0 dB) trains for one epoch over its own disjoint partition before the next begins. Together, the three stages make one pass over the complete robustness dataset. Student adapters carry over; the optimizer and schedule restart. Resuming restores the interrupted stage. Length-aware batching stays within each disjoint SNR partition. Continued-SFT targets match distillation examples by acoustic-view ID.

OPSD and OP-CAD use $\tau=1$ and complete-vocabulary distributions. OP-CAD uses $\lambda=0.5$, so weights lie in $[1,1.5]$, and $\epsilon=10^{-12}$ guards against numerical zero. Parameters do not vary by benchmark or SNR. Learning-rate warmup is separate from supervised warm-up. Each method has one training run.

\paragraph{GRPO sampling and rewards.}
GRPO samples eight responses per prompt with temperature 0.7, top-$p$ 0.95, and a 128-token limit. Correctness and format rewards have weights 1.0 and 0.2; the KL coefficient is 0.02. Batch size counts prompts, not completions.

\end{document}